\documentclass{article}

\PassOptionsToPackage{numbers}{natbib}

 \usepackage[final]{nips_2018}

\usepackage[utf8]{inputenc} 
\usepackage[T1]{fontenc}    
\usepackage{hyperref}       
\usepackage{url}            
\usepackage{booktabs}       
\usepackage{amsfonts}       
\usepackage{nicefrac}       
\usepackage{microtype}      

\usepackage{wrapfig}

\usepackage{amsthm}
\usepackage{mathtools}
\usepackage{amsmath}
\usepackage{amssymb}
\usepackage{bm}
\usepackage{upgreek}

\usepackage{algorithm}
\usepackage{algorithmic}

\usepackage{setspace}
\usepackage{xspace}
\usepackage{cancel}

\usepackage[capitalize]{cleveref}

\usepackage[disable]{todonotes}
\definecolor{blued}{RGB}{70,197,221}

\title{Statistical and Computational Trade-Offs\\* in Kernel K-Means}

\author{
    Daniele Calandriello \\
   LCSL -- IIT \& MIT, \\
    Genoa, Italy \\
   \And
   Lorenzo Rosasco \\
   University of Genoa, \\
   LCSL -- IIT \& MIT \\
}

\def\:#1{\protect \ifmmode {\mathbf{#1}} \else {\textbf{#1}} \fi}

\newcommand{\wt}[1]{\widetilde{#1}}

\newcommand{\bPhi}{\mathbf{\Upphi}}
\newcommand{\bphi}{\mathbf{\upphi}}

\newcommand{\bLambda}{\bm{\Lambda}}
\newcommand{\bPi}{\bm{\Uppi}}

\newcommand{\bc}{\mathbf{c}}

\newcommand{\be}{\mathbf{e}}
\newcommand{\bff}{\mathbf{f}}

\newcommand{\bx}{\mathbf{x}}
\newcommand{\by}{\mathbf{y}}

\newcommand{\bC}{\mathbf{C}}

\newcommand{\bF}{\mathbf{F}}

\newcommand{\bI}{\mathbf{I}}

\newcommand{\bK}{\mathbf{K}}

\newcommand{\bP}{\mathbf{P}}

\newcommand{\bR}{\mathbf{R}}
\newcommand{\bS}{\mathbf{S}}

\newcommand{\bU}{\mathbf{U}}

\newcommand{\mnistsmall}{\textsc{mnist60k}\xspace}
\newcommand{\mnistbig}{\textsc{mnist8m}\xspace}

\newcommand{\nystrom}{Nystr\"{o}m\xspace}

\newcommand{\voronoispace}{\mathcal{V}}

\newcommand{\risk}{\mathcal{E}}

\newcommand{\coldict}{\mathcal{I}}

\newcommand{\X}{\mathcal{X}} 
\newcommand{\sampdist}{\mu}

\newcommand{\pimat}{\bPi}

\newcommand{\phimat}{\bPhi}
\newcommand{\phivec}{\bphi}
\newcommand{\aphivec}{\wt{\phivec}}

\usepackage{bbm}
\newcommand{\onevec}{\mathbbm{1}}

\newtheorem{theorem}{Theorem}
\newtheorem{corollary}{Corollary}
\newtheorem{definition}{Definition}
\newtheorem{lemma}{Lemma}
\newtheorem{proposition}{Proposition}

\newcommand{\featmap}{\varphi}
\newcommand{\afeatmap}{\wt{\varphi}}

\newcounter{cnt-lem-quad-variation}
\newcommand{\feasibleset}{\mathcal{S}}

\newcommand{\wrt}{w.r.t.\@\xspace}
\newcommand{\whp}{w.h.p.\@\xspace}
\newcommand{\ie}{i.e.,\@\xspace}

\newcommand{\iid}{i.i.d.\@\xspace}

\DeclareMathOperator*{\argmin}{arg\,min}

\newcommand{\bigotime}{\mathcal{O}}
\newcommand{\bigomegatime}{\Omega}
\newcommand{\abigotime}{\wt{\mathcal{O}}}

\newcommand{\cluster}{\mathcal{C}}

\newcommand{\norm}[1]{\left\Vert #1 \right\Vert}
\newcommand{\normsmall}[1]{\Vert #1 \Vert}

\newcommand{\transp}{\mathsf{\scriptscriptstyle T}}
\DeclareMathOperator*{\Tr}{Tr}

\DeclareMathOperator*{\Ran}{Im}

\DeclareMathOperator*{\Rank}{Rank}

\DeclareMathOperator*{\expectedvalue}{\mathbb{E}}

\newcommand{\indfunc}{\mathbb{I}}

\newcommand{\Real}{\mathbb{R}}

\newcommand{\statespace}{\mathcal{X}}

\newcommand{\dataset}{\mathcal{D}}

\newcommand{\rkhs}{\mathcal{H}}

\newcommand{\kerfunc}{\mathcal{K}}
\newcommand{\kermatrix}{{\bK}}
\newcommand{\deff}{d_{\text{eff}}}

\newcommand{\akermatrix}{\mathbf{\wt{K}}}

\begin{document}

\maketitle

\begin{abstract}
We investigate the efficiency of k-means  in terms of both statistical and computational requirements.
More precisely,  we study  a Nystr\"om approach 
to kernel k-means. We analyze the statistical properties 
of the proposed method and show that it achieves  the same accuracy of 
exact kernel k-means with only a fraction of computations.
Indeed, we prove under basic assumptions  that sampling  $\sqrt{n}$ Nystr\"om  landmarks allows to greatly reduce computational costs
without incurring in any loss of accuracy. To the best of our knowledge this is the first result of this kind for unsupervised learning.
\end{abstract}

\section{Introduction}

Modern applications require machine learning algorithms to be accurate as well as computationally efficient, since data-sets are increasing in size and dimensions. Understanding the interplay and trade-offs between statistical and computational requirements is then crucial \cite{rudi2015less, rudi2017generalization}. 
In this paper,  we consider this question in the context of clustering, considering a popular nonparametric approach, namely kernel k-means \cite{scholkopf_nonlinear_1998}.
K-means is arguably one of most common approaches to clustering and produces clusters with piece-wise linear boundaries. Its kernel version allows to consider nonlinear boundaries, greatly improving the flexibility of the approach.  Its statistical properties have been studied \cite{canas_learning_2012, maurer_2010, biau2008performance} and from a computational point of view it requires manipulating an empirical kernel matrix.  As for other kernel methods, this latter operation becomes unfeasible for large scale problems and deriving approximate computations is subject of recent works, see for example \cite{tropp2017fixed, chitta_approximate_2011, rahimi2007random, wang2017scalable, musco2016provably} and reference therein. 

In this paper we are interested into quantifying the statistical effect of computational approximations.
Arguably one could expect the latter to induce some loss of accuracy. In fact, we  prove that, perhaps surprisingly, 
there  are favorable regimes where it is possible  maintain optimal statistical accuracy while significantly reducing  
computational costs. While a similar phenomenon has been recently  shown in supervised learning
\cite{rudi2015less, rudi2017generalization, calandriello2017efficient}, 
we are not aware of similar results  for other  learning tasks. 

Our approach is based on considering a Nystr\"om approach to kernel k-means based on sampling a subset of training set points
(landmarks) that can be used to approximate the kernel matrix \cite{alaoui2015fast, tropp2017fixed, calandriello_2017_icmlskons, calandriello_disqueak_2017,  wang2017scalable, musco2016provably}. 
While there is a vast literature on the properties of Nystr\"om methods for kernel approximations \cite{musco2016provably,alaoui2015fast}, experience from supervised learning show that
better results can be derived focusing on the task of interest, see discussion
in   \cite{bach2013sharp}.   The properties of Nystr\"om approximations for
k-means has been recently studied in \cite{wang2017scalable,musco2016provably}.
Here they focus only on the computational aspect of the problem, and
provide fast methods that achieve an \emph{empirical} cost only
a multiplicative factor larger than the optimal one.

Our analysis is aimed at combining both statistical {\em and} computational
results. Towards this end we derive a novel \emph{additive} bound on the
empirical cost that can be used to bound the true object of interest: the
\emph{expected cost}. This result can be combined with probabilistic results
to show that optimal statistical accuracy can be obtained considering only
$O(\sqrt{n})$ \nystrom landmark points, where $n$ is the number of training set
of points. Moreover, we show similar bounds not only for the optimal solution,
which is hard to compute in general,
but also for approximate solutions that can be computed efficiently using
$k$-means++. From a computational point of view this leads to massive
improvements reducing the memory complexity from $O(n^2)$ to $O(n\sqrt{n})$.
Experimental results  on large scale data-sets confirm and  illustrate our
findings.

The rest of the paper is organized as follows. We first overview kernel
$k$-means, and introduce our approximate kernel $k$-means approach based
on \nystrom embeddings. We then prove our statistical and computational
guarantees and empirically validate them. Finally, we present some
limits of our analysis, and open questions.

 \section{Background}
\paragraph{Notation} Given an input space $\statespace$, a sampling distribution $\sampdist$, and $n$ samples
$\{\bx_i\}_{i=1}^n$ drawn \iid
from $\sampdist$, we denote with $\sampdist_n(A) = (1/n)\sum_{i=1}^n \indfunc\{\bx_i \in A\}$
the \emph{empirical} distribution. Once the data has been sampled, we use
the \emph{feature map} $\featmap(\cdot) : \statespace \rightarrow \rkhs$
to maps $\statespace$ into a Reproducing Kernel Hilbert Space (RKHS) $\rkhs$ \cite{scholkopf2001learning}, that we assume separable,
such that for any $\bx \in \statespace$ we have $\phivec = \featmap(\bx)$. Intuitively,
in the rest of the paper the reader can assume that $\phivec \in \Real^D$ with $D \gg n$ or even infinite.
Using the kernel trick \cite{Aizerman67theoretical}
we also know that $\phivec^\transp\phivec' = \kerfunc(\bx, \bx')$,
where $\kerfunc$ is the kernel function associated with $\rkhs$ and
$\phivec^\transp\phivec' = \langle\phivec, \phivec'\rangle_{\rkhs}$
is a short-hand for the inner product in $\rkhs$. With a slight abuse of
notation we will also define the norm $\norm{\phivec}^2 = \phivec^\transp\phivec$,
and assume that $\norm{\featmap(\bx)}^2 \leq \kappa^2$ for any $\bx \in \statespace$.
Using $\phivec_i = \featmap(\bx_i)$, we denote with $\dataset = \{\phivec_i\}_{i=1}^n$ the input dataset. We also represent the dataset as the map $\phimat_n = [\phivec_1, \dots, \phivec_n] : \Real^n \rightarrow \rkhs$
with $\phivec_i$ as its $i$-th column.
We denote with $\kermatrix_{n} = \phimat_n^\transp\phimat_n$ the empirical
kernel matrix with entries $[\kermatrix_n]_{i,j} = k_{i,j}$.
Finally, given $\phimat_n$
we denote as $\pimat_n = \phimat_n\phimat_n^\transp(\phimat_n\phimat_n^\transp)^{+}$
the orthogonal projection matrix on the span $\rkhs_n = \Ran(\phimat_n)$
of the dataset.

\paragraph{$k$-mean's objective} Given our dataset, we are interested in partitioning it into $k$ disjoint \emph{clusters}
each characterized by its \emph{centroid} $\bc_j$. The Voronoi cell
associated with a centroid $\bc_j$ is defined as the set
$\cluster_j := \{i : j = \argmin_{s = [k]} \norm{\phivec_i - \bc_s}^2\}$,
or in other words a point $\phivec_i \in \dataset$ belongs to the $j$-th
cluster if $\bc_j$ is its closest centroid.
Let $\bC = [\bc_1, \dots \bc_k]$ be a collection of $k$ centroids from $\rkhs$.
We can now formalize the criterion we use to measure clustering quality.
\begin{definition}\label{def:object-Cn}
The empirical and expected squared norm criterion are defined as
\begin{align*}
    W(\bC, \mu_n) := \frac{1}{n}\sum_{i=1}^n\min_{j=1,\dots,k}\normsmall{\phivec_i - \bc_j}^2,
    &&
    W(\bC, \mu) := \expectedvalue_{\phivec \sim \mu}\left[ \min_{j=1,\dots,k}\normsmall{\phivec - \bc_j}^2 \right].
\end{align*}
The empirical risk minimizer (ERM) is defined as $\bC_n := \argmin_{\bC \in \rkhs^{k}} W(\bC, \mu_n)$.
\end{definition}
The sub-script $n$ in $\bC_n$ indicates that it minimizes $W(\bC, \mu_n)$ for the $n$ samples in $\dataset$.
\citet{biau2008performance} gives us a bound on the excess risk of the empirical risk minimizer.
\begin{proposition}[\cite{biau2008performance}]\label{prop:biau-risk-bound}
The excess risk $\risk(\bC_n)$ of the empirical risk minimizer $\bC_n$ satisfies
\begin{align*}
\risk(\bC_n) := \expectedvalue\nolimits_{\dataset \sim \sampdist} \left[W(\bC_n, \mu)\right] - W^{*}(\mu) \leq \bigotime\left(k/\sqrt{n}\right)
\end{align*}
where $W^*(\mu) := \inf_{\bC \in \rkhs^{k}} W(\bC, \mu)$ is the optimal clustering risk.
\end{proposition}
From a theoretical perspective, this result is only $\sqrt{k}$ times larger than a corresponding $\bigotime(\sqrt{k/n})$
lower bound \cite{graf_foundations_2000}, and therefore shows that the ERM $\bC_n$
achieve an excess risk optimal in $n$.
From a computational perspective, \cref{def:object-Cn} cannot be directly
used to compute $\bC_n$, since the points $\phivec_i$
in $\rkhs$ cannot be directly represented.
Nonetheless, due to properties of the squared norm criterion, each
$\bc_j \in \bC_n$ must be the mean of all $\phivec_i$ associated
with that center, \ie $\bC_n$ belongs to $\rkhs_n$. Therefore, it can be explicitly represented
as a sum of the $\phivec_i$ points included in the $j$-th cluster, \ie all the points in the $j$-th Voronoi cell $\cluster_j$. Let $\voronoispace$ be the space of all possible disjoint partitions $[\cluster_1,\dots,\cluster_j]$. We can use this fact, together with the kernel trick, to reformulate
the objective $W(\cdot, \mu_n)$.

\begin{proposition}[\cite{dhillon_unified_2004}]\label{prop:full-hn-embedding}
We
can rewrite the objective
\begin{align*}
\min_{\bC \in \rkhs}  W(\bC, \mu_n)
    &= \frac{1}{n}\min_{\voronoispace} \sum_{j=1}^k\sum_{i \in \cluster_j}\Bigg\Vert{\phivec_i - \frac{1}{|\cluster_j|}\sum_{s \in \cluster_j}\phivec_{s}}\Bigg\Vert^2
\end{align*}
with 
$
\norm{\phivec_i - \frac{1}{|\cluster_j|}\phivec_{\cluster_j}\onevec_{|\cluster_j|}}^2
= k_{i,i} - \frac{2}{|\cluster_j|}\sum_{s \in \cluster_j}k_{i,s} + \frac{1}{|\cluster_j|^2}\sum_{s \in \cluster_j}\sum_{s' \in \cluster_j}k_{s,s'}
$
\end{proposition}
While the combinatorial search over $\voronoispace$ can now
be explicitly computed and optimized using the kernel matrix $\kermatrix_n$, it still remains highly inefficient to do
so. In particular, simply constructing and storing $\kermatrix_n$
takes $\bigotime(n^2)$ time and space and does not scale to large datasets.

 \section{Algorithm}
A simple approach to
reduce computational cost is to use approximate embeddings, which replace the map $\featmap(\cdot)$ and points $\phivec_i = \featmap(\bx_i) \in \rkhs$ with a finite-dimensional approximation
$\aphivec_i = \wt{\featmap}(\bx_i) \in \Real^m$.

\paragraph{\nystrom kernel $k$-means}

Given a dataset $\dataset$, we denote with $\coldict = \{\phivec_j\}_{j=1}^m$
a \emph{dictionary} (\ie subset) of $m$ points $\phivec_j$ from $\dataset$,
and with $\phimat_m: \Real^m \rightarrow \rkhs$ the map with these points as columns. These points acts as landmarks \cite{williams2001using}, inducing
a smaller space $\rkhs_m = \Ran(\phimat_m)$ spanned by the dictionary.
As we will see in the next section, $\coldict$ should be chosen so that $\rkhs_m$ is close to the whole span $\rkhs_n = \Ran(\phimat_n)$ of the dataset.

Let $\kermatrix_{m,m} \in \Real^{m \times m}$
be the empirical kernel matrix between all points in $\coldict$, and denote with
\begin{align}\label{eq:def-pimat-dict}
\pimat_{m}
= \phimat_m\phimat_m^\transp(\phimat_m\phimat_m^\transp)^{+}
= \phimat_m\kermatrix_{m,m}^{+}\phimat_m^\transp,
\end{align}
the orthogonal projection on $\rkhs_m$.
Then we can define
an approximate ERM over $\rkhs_m$ as
\begin{align}\label{eq:c_nm_bad_def}
\bC_{n,m}
= \argmin_{\bC \in \rkhs_{m}^k} \frac{1}{n}\sum_{i=1}^n\min_{j=[k]}\normsmall{\phivec_i - \bc_j}^2
= \argmin_{\bC \in \rkhs_{m}^k} \frac{1}{n}\sum_{i=1}^n\min_{j=[k]}\normsmall{\pimat_m(\phivec_i - \bc_j)}^2,
\end{align}
since any component outside of $\rkhs_m$ is just a constant in the minimization.
Note that the centroids $\bC_{n,m}$ are still points in $\rkhs_m \subset \rkhs$, and we cannot directly
compute them.
Instead, we can use the eigen-decomposition of $\kermatrix_{m,m} = \bU\bLambda\bU^\transp$
to rewrite $\pimat_m = \phimat_{m}\bU\bLambda^{-1/2}\bLambda^{-1/2}\bU^\transp\phimat_{m}^\transp$.
Defining now $\wt{\featmap}(\cdot) = \bLambda^{-1/2}\bU^\transp\phimat_{m}^\transp\featmap(\cdot)$
we have a finite-rank embedding into $\Real^m$.
Substituting in \cref{eq:c_nm_bad_def}
\begin{align*}
\normsmall{\pimat_m(\phivec_i - \bc_j)}^2
=\normsmall{\bLambda^{-1/2}\bU^\transp\phimat_{m}^\transp(\phivec_i - \bc_j)}^2
=\normsmall{\aphivec_i - \bLambda^{-1/2}\bU^\transp\phimat_{m}^\transp\bc_j}^2,
\end{align*}
where $\aphivec_i := \bLambda^{-1/2}\bU^\transp\phimat_{m}^\transp\phivec_i$ are the embedded points.
Replacing $\wt{\bc}_j := \bLambda^{-1/2}\bU^\transp\phimat_{m}^\transp\bc_j$ and searching over $\wt{\bC} \in \Real^{m \times k}$
instead of searching over $\bC \in \rkhs_m^k$, we obtain (similarly to \cref{prop:full-hn-embedding})
\begin{align}\label{eq:def-approx-optim}
\wt{\bC}_{n,m} = \argmin_{\wt{\bC} \in \Real^{m \times k}} \frac{1}{n}\sum_{i=1}^n\min_{j=[k]}\normsmall{\aphivec_i - \wt{\bc}_j}^2
= \frac{1}{n}\min_{\voronoispace} \sum_{j=1}^k\sum_{i \in \cluster_j}\norm{\aphivec_i - \frac{1}{|\cluster_j|}\sum_{s \in \cluster_j}\aphivec_s}^2,
\end{align}
where we do not need to resort to kernel tricks, but can use the $m$-dimensional embeddings $\aphivec_i$
to explicitly compute the centroid $\sum_{s \in \cluster_j}\aphivec_s$.
\cref{eq:def-approx-optim} can now be solved in
multiple ways. The most straightforward is to run a parametric $k$-means algorithm
to compute $\wt{\bC}_{n,m}$, and then invert the relationship $\wt{\bc}_j = \phimat_{m}\bU\bLambda^{-1/2}\bc_j$ to bring
back the solution to $\rkhs_m$, \ie $\bC_{n,m} = \phimat_m^+\bU^\transp\bLambda^{1/2}\wt{\bC}_{n,m} = \phimat_m\bU\bLambda^{-1/2}\wt{\bC}_{n,m}$.
This can be done in in $\bigotime(nm)$ space and $\bigotime(nmkt + nm^2)$ time
using $t$ steps of Lloyd's algorithm \cite{lloyd1982least} for $k$ clusters.
More in detail, computing the embeddings $\aphivec_i$ is a one-off cost taking $nm^2$ time.
Once the $m$-rank \nystrom embeddings $\aphivec_i$ are computed they can be
stored and manipulated in $nm$ time and space, with an $n/m$ improvement over the $n^2$ time and
space required to construct $\kermatrix_n$.

\begin{algorithm}[t]
\begin{algorithmic}
\REQUIRE dataset $\dataset = \{\phivec_i\}_{i=1}^n$, dictionary $\coldict = \{\phivec_j\}_{j=1}^m$ with points from $\dataset$, number of clusters $k$
\STATE compute kernel matrix $\kermatrix_{m,m} = \phimat_{m}^\transp\phimat_{m}$ between all points in $\coldict$
\STATE compute eigenvectors $\bU$ and eigenvalues $\bLambda$ of $\kermatrix_{m,m}$
\STATE for each point $\phivec_i$, compute embedding $\aphivec_i = \bLambda^{-1/2}\bU^\transp\phimat_{m}^\transp\phivec_i = \bLambda^{-1/2}\bU^\transp\kermatrix_{m,i} \in \Real^{m}$
\STATE compute optimal centroids $\wt{\bC}_{n,m} \in \Real^{m \times k}$ on the embedded dataset $\wt{\dataset} = \{\aphivec_i\}_{i=1}^n$
\STATE compute explicit representation of centroids $\bC_{n,m} = \phimat_{m}\bU\bLambda^{-1/2}\wt{\bC}_{n,m}$
\end{algorithmic}
\caption{\nystrom Kernel K-Means}\label{alg:nyst-kk-means}
\end{algorithm}

 \subsection{Uniform sampling for dictionary construction}
Due to its derivation, the computational cost of \cref{alg:nyst-kk-means} depends on the size $m$
of the dictionary $\coldict$. Therefore, for computational reasons we would prefer
to select as small a dictionary as possible.
As a conflicting goal, we also wish to optimize $W(\cdot, \mu_n)$ well,
which requires a $\wt{\featmap}(\cdot)$ and $\coldict$ rich
enough to approximate $W(\cdot, \mu_n)$ well.
Let $\pimat_m^{\bot}$ be the projection orthogonal to $\rkhs_m$. Then when
$\bc_i \in \rkhs_m$

\vspace{-.5\baselineskip}
\begin{align*}
\normsmall{\phivec_i - \bc_i}^2 = \normsmall{(\pimat_m + \pimat_m^\bot)(\phivec_i - \bc_i)}^2 = \normsmall{\pimat_m(\phivec_i - \bc_i)}^2 + \normsmall{\pimat_m^{\bot}\phivec_i}^2.
\end{align*}

We will now introduce the concept of a $\gamma$-preserving dictionary $\coldict$
to control the quantity $\normsmall{\pimat_m^{\bot}\phivec_i}^2$.
\begin{definition}\label{def:regularized-nyst-app-guar}
We define the subspace $\rkhs_m$ and dictionary $\coldict$ as $\gamma$-preserving
\wrt space $\rkhs_n$ if

\vspace{-\baselineskip}
\begin{align}
\pimat_m^{\bot} = \pimat_n - \pimat_m \preceq \frac{\gamma}{1-\varepsilon}\left(\phimat_n\phimat_n^\transp+ \gamma\pimat_n\right)^{-1}.
\end{align}
\vspace{-.75\baselineskip}
\end{definition}
Notice that the inverse $\left(\phimat_n\phimat_n^\transp + \gamma\pimat_n\right)^{-1}$ on the
right-hand side of the inequality is crucial to control
the error $\normsmall{\pimat_m^{\bot}\phivec_i}^2 \lesssim \gamma \phivec_i^\transp\left(\phimat_n\phimat_n^\transp + \gamma\pimat_n\right)^{-1}\phivec_i$.
In particular, since $\phivec_i \in \phimat_n$, we have that in the \emph{worst case}
the error is bounded as $\phivec_i^\transp\left(\phimat_n\phimat_n^\transp + \gamma\pimat_n\right)^{-1}\phivec_i
\leq \phivec_i^\transp\left(\phivec_i\phivec_i^\transp\right)^{+}\phivec_i \leq 1$.
Conversely, since $\lambda_{\max}(\phimat_n\phimat_n^\transp) \leq \kappa^2n$ we know that in the best case the error can be reduced up to
$1/n \leq \phivec_i^\transp\phivec_i/\lambda_{\max}(\phimat_n\phimat_n^\transp) \leq \phivec_i^\transp\left(\phimat_n\phimat_n^\transp + \gamma\pimat_n\right)^{-1}\phivec_i$.
Note that the directions associated with the larger eigenvalues are the ones
that occur most frequently in the data. As a consequence, \cref{def:regularized-nyst-app-guar}
guarantees that the overall error across the whole dataset remains small.
In particular, we can control the residual $\pimat_m^{\bot}\phimat_n$ after the projection as $\normsmall{\pimat_m^{\bot}\phimat_n}^2 \leq \gamma\normsmall{\phimat_n^\transp(\phimat_n\phimat_n^\transp + \gamma\pimat_n)^{-1}\phimat_n} \leq \gamma$.

To construct $\gamma$-preserving dictionaries we focus
on a uniform random sampling approach\cite{bach2013sharp}.
Uniform sampling is historically the first \cite{williams2001using}, and usually the simplest approach
used to construct $\coldict$.
Leveraging results from the literature \cite{bach2013sharp,calandriello_disqueak_2017,musco2016provably}
we can show that 
uniformly sampling $\abigotime(n/\gamma)$ landmarks
generates a $\gamma$-preserving dictionary with high probability.
\begin{lemma}\label{lem:uniform-sampling-preserving}
For a given $\gamma$, construct $\coldict$ by uniformly
sampling $m \geq 12\kappa^2n/\gamma\log(n/\delta)/\varepsilon^2$
landmarks from $\dataset$.
Then w.p.\@ at least $1 - \delta$
the dictionary $\coldict$ is $\gamma$-preserving.
\end{lemma}
\citet{musco2016provably} obtains a similar result, but instead of considering
the operator $\pimat_n$ they focus on the finite-dimensional eigenvectors of $\kermatrix_n$.
Moreover, their $\pimat_n \preceq 
 \pimat_m + \frac{\varepsilon\gamma}{1-\varepsilon}\left(\phimat_n\phimat_n^\transp\right)^{+}$ bound 
is weaker and would not be sufficient to satisfy our definition of $\gamma$-accuracy.
A result equivalent to \cref{lem:uniform-sampling-preserving} was obtained by \citet{alaoui2015fast},
but they also only focus on the finite-dimensional eigenvectors of $\kermatrix_n$,
and did not investigate the implications for $\rkhs$.
\begin{proof}[Proof sketch of \cref{lem:uniform-sampling-preserving}]
It is well known \cite{bach2013sharp,calandriello_disqueak_2017} that uniformly
sampling $\bigotime(n/\gamma\varepsilon^{-2}\log(n/\delta))$ points
with replacement is sufficient to obtain w.p.~$1-\delta$
the following guarantees on $\phimat_m$
\begin{align*}
(1-\varepsilon)\phimat_n\phimat_n^\transp - \varepsilon\gamma\pimat_n \preceq
\frac{n}{m}\phimat_m\phimat_m^\transp\preceq
(1+\varepsilon)\phimat_n\phimat_n^\transp + \varepsilon\gamma\pimat_n.
\end{align*}
Which implies
\begin{align*}
\left(\frac{n}{m}\phimat_m\phimat_m^\transp + \gamma\pimat_n\right)^{-1}
\preceq \left((1-\varepsilon)\phimat_n\phimat_n^\transp - \varepsilon\gamma\pimat_n + \gamma\pimat_n\right)^{-1}
= \frac{1}{1-\varepsilon}\left(\phimat_n\phimat_n^\transp + \gamma\pimat_n\right)^{-1}
\end{align*}

We can now rewrite $\pimat_n$ as

\vspace{-\baselineskip}
\begin{align*}
\pimat_n &= \left(\frac{n}{m}\phimat_m\phimat_m^\transp + \gamma\pimat_n\right)\left(\frac{n}{m}\phimat_m\phimat_m^\transp + \gamma\pimat_n\right)^{-1}\\
&= \frac{n}{m}\phimat_m\phimat_m^\transp\left(\frac{n}{m}\phimat_m\phimat_m^\transp + \gamma\pimat_n\right)^{-1} + \gamma\left(\frac{n}{m}\phimat_m\phimat_m^\transp + \gamma\pimat_n\right)^{-1}\\
&\preceq \frac{n}{m}\phimat_m\phimat_m^\transp\left(\frac{n}{m}\phimat_m\phimat_m^\transp\right)^{+} + \gamma\left(\frac{n}{m}\phimat_m\phimat_m^\transp + \gamma\pimat_n\right)^{-1}\\
&= \pimat_m + \gamma\left(\frac{n}{m}\phimat_m\phimat_m^\transp + \gamma\pimat_n\right)^{-1}
\preceq \pimat_m + \frac{\gamma}{1-\varepsilon}\left(\phimat_n\phimat_n^\transp + \gamma\pimat_n\right)^{-1}
\end{align*}

\vspace{-1.5\baselineskip}
\end{proof}
In other words, using uniform sampling
we can reduce the size of the search space $\rkhs_m$ by a $1/\gamma$ factor
(from $n$ to $m \simeq n/\gamma$) in exchange for a $\gamma$ additive error,
resulting in a computation/approximation trade-off that
is linear in $\gamma$.

 \section{Theoretical analysis}
Exploiting the error bound for $\gamma$-preserving dictionaries we are now
ready for the main result of this paper: showing that we can improve
the computational aspect of kernel $k$-means using \nystrom embedding,
while maintaining optimal generalization guarantees.
\begin{theorem}\label{thm:main-result}
Given a $\gamma$-preserving dictionary
\begin{align*}
\risk(\bC_{n,m}) = W(\bC_{n,m}, \mu) - W(\bC_{n}, \mu) \leq \bigotime\left(k\left(\frac{1}{\sqrt{n}} + \frac{\gamma}{n}\right)\right)
\end{align*}
\end{theorem}
From a statistical point of view, \cref{thm:main-result}
shows that if $\coldict$ is $\gamma$-preserving, the ERM
in $\rkhs_m$ achieves the same \emph{excess} risk as the
exact ERM from $\rkhs_n$ up to an additional $\gamma/n$ error.
Therefore, choosing $\gamma = \sqrt{n}$
the solution $\bC_{n,m}$ achieves the
$\bigotime\left(k/\sqrt{n}\right) + \bigotime(k\sqrt{n}/n) \leq \bigotime(k/\sqrt{n})$
generalization \cite{biau2008performance}.
Our results can also be applied to traditional $k$-means in Euclidean space,
i.e.~the special case of $\rkhs = \Real^d$ and a linear kernel.
In this case \cref{thm:main-result} shows that regardless of the size of the input space $d$,
it is possible to embed all points in $\Real^{\sqrt{n}}$ and preserve optimal statistical
rates. In other words, given $n$ samples we can always construct $\sqrt{n}$ features
that are sufficient to well preserve the objective regardless of the original number of features $d$.
In the case when $d \gg n$, this can lead to substantial computational improvements.

From a computational point of view, \cref{lem:uniform-sampling-preserving}
shows that we can construct an $\sqrt{n}$-preserving dictionary
simply by sampling $\abigotime(\sqrt{n})$ points uniformly\footnote{$\abigotime$ hides logarithmic dependencies on $n$ and $m$.},
which greatly reduces the embedding size from $n$ to $\sqrt{n}$, and the total
required space from $n^2$ to $\abigotime(n\sqrt{n})$.

Time-wise, the bottleneck becomes
the construction of the embeddings $\aphivec_i$, which takes $nm^2 \leq \abigotime(n^2)$
time, while each iterations of Lloyd's algorithm only requires $nm \leq \abigotime(n\sqrt{n})$ time.
In the full generality of our setting this is practically optimal, since computing
a $\sqrt{n}$-preserving dictionary is in general as hard as matrix multiplication
\cite{NIPS2017_7030,backurs2017fine}, which requires $\bigomegatime(n^2)$ time.
In other words, unlike the case of space complexity, there is no free
lunch for time complexity, that in the worst case must scale as $n^2$ similarly to
the exact case.
Nonetheless embedding the points is an embarrassingly parallel problem that can
be easily distributed, while in practice it is usually the execution of the Lloyd's
algorithm that dominates the runtime.

Finally, when the dataset satisfies certain regularity conditions,
the size of $\coldict$ can be improved, which reduces both embedding and clustering runtime.
Denote with $\deff^n(\gamma)= \Tr\left(\kermatrix_n^\transp(\kermatrix_n + \bI_n)^{-1}\right)$ the so-called \emph{effective} dimension \cite{alaoui2015fast}
of $\kermatrix_n$. Since $\Tr\left(\kermatrix_n^\transp(\kermatrix_n + \bI_n)^{-1}\right) \leq \Tr\left(\kermatrix_n^\transp(\kermatrix_n)^{+}\right)$, we have that $\deff^n(\gamma) \leq r := \Rank(\kermatrix_n)$, and therefore $\deff^n(\gamma)$ can  be seen as a soft version of the rank.
When $\deff^n(\gamma) \ll \sqrt{n}$ it is possible to construct
a $\gamma$-preserving dictionary with only $\deff^n(\gamma)$ landmarks in $\abigotime(n\deff^n(\gamma)^2)$ time
using specialized algorithms \cite{calandriello_disqueak_2017} (see \cref{sec:open-quest}).
In this case, the embedding step would require only $\abigotime(n\deff^n(\gamma)^2) \ll \abigotime(n^2)$,
improving both time and space complexity.

Morever, to the best of our knowledge, this is the first
example of an unsupervised non-parametric problem where it is always (\ie without assumptions
on $\sampdist$) possible to preserve the optimal $\bigotime(1/\sqrt{n})$ risk rate
while reducing the search from the whole space $\rkhs$ to a smaller
$\rkhs_m$ subspace.

\begin{proof}[Proof sketch of \cref{thm:main-result}]
We can separate the distance between $W(\bC_{n,m}, \mu) - W(\bC_{n}, \mu)$
in a component that depends on how close $\mu$ is to $\mu_n$, bounded
using \cref{prop:biau-risk-bound}, and a component 
$W(\bC_{n,m}, \mu_n) - W(\bC_{n}, \mu_n)$ that depends on the distance
between $\rkhs_n$ and $\rkhs_m$
\begin{lemma}\label{lem:main-lemma}
Given a $\gamma$-preserving dictionary
\begin{align*}
W(\bC_{n,m}, \mu_n) - W(\bC_{n}, \mu_n) \leq \frac{\min(k,\deff^n(\gamma))}{1-\varepsilon}\frac{\gamma}{n}
\end{align*}
\end{lemma}
To show this we can rewrite
the objective as (see \cite{dhillon_unified_2004})
\begin{align*}
W(\bC_{n,m}, \mu_n) = 
\normsmall{\phimat_n - \pimat_m\phimat_n\bS_{n,m}}_F^2
= \Tr(\phimat_n^\transp\phimat_n - \bS_{n}\phimat_n^\transp\pimat_m\phimat_n\bS_{n}),
\end{align*}
where $\bS_n \in \Real^{n \times n}$ is a $k$-rank \emph{projection} matrix associated
with the \emph{exact} clustering $\bC_n$. Then using \cref{def:regularized-nyst-app-guar}
we have $\pimat_m - \pimat_n \succeq  -\frac{\gamma}{1-\varepsilon}(\phimat_n\phimat_n^\transp + \gamma\pimat_n)^{-1}$ and
we obtain an \emph{additive} error bound
\begin{align*}
&\Tr(\phimat_n^\transp\phimat_n - \bS_{n}\phimat_n^\transp\pimat_m\phimat_n\bS_{n})\\
&\leq \Tr\left(\phimat_n^\transp\phimat_n - \bS_{n}\phimat_n^\transp\phimat_n\bS_{n}
+ \frac{\gamma}{1-\varepsilon}\bS_{n}\phimat_n^\transp(\phimat_n\phimat_n^\transp + \gamma\pimat_n)^{-1}\phimat_n\bS_{n}\right)\\
&= W(\bC_n,\sampdist_n)
+ \frac{\gamma}{1-\varepsilon}\Tr\left(\bS_{n}\phimat_n^\transp(\phimat_n\phimat_n^\transp + \gamma\pimat_n)^{-1}\phimat_n\bS_{n}\right).
\end{align*}

Since $\normsmall{\phimat_n^\transp(\phimat_n\phimat_n^\transp + \gamma\pimat_n)^{-1}\phimat_n} \leq 1$,
$\bS_n$ is a projection matrix, and $\Tr(\bS_{n}) = k$ we have
\begin{align*}
\tfrac{\gamma}{1-\varepsilon}\Tr\left(\bS_{n}\phimat_n^\transp(\phimat_n\phimat_n^\transp + \gamma\pimat_n)^{-1}\phimat_n\bS_{n}\right)
\leq \tfrac{\gamma}{1-\varepsilon}\Tr\left(\bS_{n}\bS_{n}\right)
=  \tfrac{\gamma k}{1-\varepsilon}.
\end{align*}
Conversely, if we focus on the matrix $\phimat_n^\transp(\phimat_n\phimat_n^\transp + \gamma\pimat_n)^{-1}\phimat_n \preceq \pimat_n$ we have
\begin{align*}
\tfrac{\gamma}{1-\varepsilon}\Tr\left(\bS_{n}\phimat_n^\transp(\phimat_n\phimat_n^\transp + \pimat_n)^{-1}\phimat_n\bS_{n}\right)
\leq \tfrac{\gamma}{1-\varepsilon}\Tr\left(\phimat_n^\transp(\phimat_n\phimat_n^\transp + \pimat_n)^{-1}\phimat_n\right)
\leq  \tfrac{\gamma \deff^n(\gamma)}{1-\varepsilon}.
\end{align*}
Since both bounds hold simultaneously, we can simply take the minimum to conclude
our proof.
\end{proof}

We now compare the theorem with previous work. Many approximate kernel $k$-means methods have been proposed over the years, and can be roughly split in two groups.

Low-rank \emph{decomposition} based methods try to directly
simplify the optimization problem from \cref{prop:full-hn-embedding}, replacing the kernel
matrix $\kermatrix_n$ with an approximate $\akermatrix_n$ that can be stored and
manipulated more efficiently. Among these methods we can mention partial decompositions
\cite{bach2005predictive}, \nystrom approximations based on uniform \cite{williams2001using}, $k$-means++ \cite{oglic2017nystrom},
or ridge leverage score (RLS) sampling\cite{wang2017scalable, musco2016provably, calandriello_disqueak_2017}, and
random-feature approximations \cite{pmlr-v70-avron17a}.
None of these optimization based methods focus on the underlying excess risk problem,
and their analysis cannot be easily integrated in existing results,
as the approximate minimum found has no clear interpretation as a statistical
ERM.

Other works take the same \emph{embedding} approach that we do, and directly replace
the exact $\featmap(\cdot)$ with an approximate $\afeatmap(\cdot)$,
such as \nystrom embeddings \cite{williams2001using}, Gaussian projections \cite{biau2008performance},
and again random-feature approximations \cite{rahimi2007random}. Note that these approaches also
result in approximate $\akermatrix_n$ that can be manipulated efficiently,
but are simpler to analyze theoretically.
Unfortunately, no existing embedding based methods can guarantee at the
same time optimal excess risk rates and a reduction in the size of
$\rkhs_m$, and therefore a reduction in computational cost.

To the best of our knowledge, the only other result providing excess risk guarantee for
approximate kernel $k$-means is \citet{biau2008performance}, where the authors
consider the excess risk of the ERM when the approximate $\rkhs_m$ is obtained using Gaussian projections.
\citet{biau2008performance} notes that the feature map
$\featmap(\bx) = \sum_{s=1}^{D} \psi_s(\bx)$ can be expressed
using an expansion of basis functions $\psi_s(\bx)$, with $D$ very large or infinite.
Given a matrix $\bP \in \Real^{m \times D}$ where each entry is a standard Gaussian r.v.,
\cite{biau2008performance} proposes the following $m$-dimensional approximate feature map
$\wt{\featmap}(\bx) = \bP[\psi_1(\bx), \dots , \psi_D(\bx)] \in \Real^m$.
Using Johnson-Lindenstrauss (JL) lemma \cite{johnson1984extensions},
they show that if $m \geq \log(n)/\nu^2$ then a multiplicative error bound of the form
$W(\bC_{n,m}, \mu_n) \leq (1+\nu)W(\bC_{n}, \mu_n)$ holds.
Reformulating their bound, we obtain that $W(\bC_{n,m}, \mu_n) - W(\bC_{n}, \mu_n) \leq \nu W(\bC_{n}, \mu_n)
\leq \nu\kappa^2$ and  $\risk(\bC_{n,m}) \leq \bigotime(k/\sqrt{n} + \nu)$.

Note that to obtain a bound comparable to \cref{thm:main-result},
and if we treat $k$ as a constant, we need to take $\nu = \gamma/n$ which results in $m \geq (n/\gamma)^2$.
This is always worse than our $\abigotime(n/\gamma)$ result for
uniform \nystrom embedding. In particular, in the $1/\sqrt{n}$ risk
rate setting Gaussian projections would require $\nu = 1/\sqrt{n}$ resulting
in $m \geq n\log(n)$ random features, which would not bring any
improvement over computing $\kermatrix_n$. Moreover when $D$ is infinite, as it is usually the
case in the non-parametric setting, the JL projection is not explicitly computable
in general and \citet{biau2008performance} must assume the existence of
a computational oracle capable of constructing $\wt{\featmap}(\cdot)$.
Finally
note that, under the hood, traditional embedding methods such as those based on 
JL lemma, usually provide only bounds of the form $\pimat_n - \pimat_m \preceq \gamma\pimat_n$,
and an error $\normsmall{\pimat_m^{\bot}\phivec_i}^2 \leq \gamma\norm{\phivec_i}^2$
(see the discussion of \cref{def:regularized-nyst-app-guar}).
Therefore the error can be larger along multiple directions,
and the overall error $\normsmall{\pimat_m^{\bot}\phimat_n}^2$ across the dictionary
can be as large as $n\gamma$ rather than $\gamma$.

Recent work in RLS sampling has also focused on bounding the distance
$W(\bC_{n,m}, \mu_n) - W(\bC_{n}, \mu_n)$ between empirical errors.
\citet{wang2017scalable} and \citet{musco2016provably} provide multiplicative
error bounds of the form $W(\bC_{n,m}, \mu_n) \leq (1+\nu)W(\bC_{n}, \mu_n)$
for uniform and RLS sampling. Nonetheless, they only focus on empirical risk and do not investigate the interaction between approximation and generalization, \ie statistics and computations.
Moreover, as we already remarked for \cite{biau2008performance}, to achieve the $1/\sqrt{n}$ excess risk rate
using a multiplicative error bound we would require an unreasonably small $\nu$,
resulting in a large $m$ that brings no computational improvement over the exact
solution.

Finally, note that when \cite{rudi2015less} showed that a favourable trade-off
was possible for kernel ridge regression (KRR), they strongly leveraged the fact
that KRR is a $\gamma$-regularized problem. Therefore, all eigenvalues and eigenvectors
in the $\phimat_n\phimat_n^\transp$ covariance matrix smaller than the $\gamma$
regularization do not influence significantly the solution.
Here we show the same for kernel $k$-means, a problem without regularization.
This hints at a deeper geometric motivation which might be at the root of both
problems, and potentially similar approaches could be leveraged in other domains.

\subsection{Further results: beyond ERM}
So far we provided guarantees for $\bC_{n,m}$, that this the ERM in $\rkhs_m$.
Although $\rkhs_m$ is much smaller than $\rkhs_n$, solving the optimization problem to find the ERM is still
NP-Hard in general \cite{aloise2009np}. Nonetheless, Lloyd's algorithm \cite{lloyd1982least}, when coupled with
a careful $k$-means++ seeding, can return a good approximate solution
$\bC_{n,m}^{++}$.
\begin{proposition}[\cite{arthur_k_meanspp_2007}]\label{prop:kmpp-opt-lemma}
For any dataset
\begin{align*}
\expectedvalue_{\mathcal{A}}[W(\bC_{n,m}^{++}, \mu_n)] \leq 8(\log(k)+ 2)W(\bC_{n,m}, \mu_n),
    \end{align*}
where $\mathcal{A}$ is the randomness deriving from the $k$-means++ initialization.
\end{proposition}
Note that, similarly to \cite{wang2017scalable, musco2016provably},
this is a multiplicative error bound on the empirical risk,
and as we discussed we cannot leverage \cref{lem:main-lemma} to bound the excess risk $\risk(\bC_{n,m}^{++})$.
Nonetheless we can still leverage \cref{lem:main-lemma}
to bound only the expected risk $W(\bC_{n,m}^{++}, \mu)$,
albeit with an extra error term appearing
that scales with the optimal clustering risk $W^*(\mu)$ (see \cref{prop:biau-risk-bound}).
\begin{theorem}\label{thm:main-kpp-result}
Given a $\gamma$-preserving dictionary
\begin{align*}
\expectedvalue_{\dataset \sim \sampdist}\left[\expectedvalue_{\mathcal{A}}[W(\bC_{n,m}^{++}, \mu)]\right] \leq \bigotime\left(\log(k)\left(\frac{k}{\sqrt{n}} + k\frac{\gamma}{n} + W^*(\mu)\right)\right).
\end{align*}
\end{theorem}
From a statistical perspective, we can once again,
set $\gamma = \sqrt{n}$ to obtain a $\bigotime(k/\sqrt{n})$
rate for the first part of the bound. Conversely, the optimal clustering risk $W^*(\mu)$
is a $\sampdist$-dependent quantity that cannot in general be bounded in $n$,
and captures how well our model, \ie the choice of $\rkhs$ and how well the criterion $W(\cdot,\mu)$,
matches reality.

From a computational perspective, we can now bound the computational
cost of finding $\bC_{n,m}^{++}$. In particular, each iteration of Lloyd's
algorithm will take only $\abigotime(n\sqrt{n}k)$ time.
Moreover, when $k$-means++ initialization is used,
the expected number of iterations required for Lloyd's algorithm to converge
is only logarithmic \cite{ailon2009streaming}.
Therefore, ignoring the time required to embed the points, we can find a solution in $\abigotime(n\sqrt{n}k)$ time and space instead of the $\abigotime(n^2k)$
cost required by the exact method, with a strong $\bigotime(\sqrt{n})$ improvement.

Finally, if the data distribution satisfies some regularity assumption the following result follows \cite{canas_learning_2012}.
\begin{corollary}
If  we denote by  $\X_\mu$ the support of the distribution $\mu$ and assume 
$\varphi(\X_\mu)$ to be a $d$-dimensional manifold,  then  $W^*(\mu) \leq dk^{-2/d}$,
and  given a $\sqrt{n}$-preserving dictionary the expected cost satisfies
\begin{align*}
\expectedvalue_{\dataset \sim \sampdist}[\expectedvalue_{\mathcal{A}}[W(\bC_{n,m}^{++}, \mu)]] \leq \bigotime\left(\log(k)\left(\frac{k}{\sqrt{n}} + dk^{-2/d}\right)\right).
\end{align*}
\end{corollary}

 \section{Experiments}
We now evaluate experimentally the claims of \cref{thm:main-result},
namely that sampling $\abigotime(n/\gamma)$ increases the excess risk by an extra
$\gamma/n$ factor, and that $m = \sqrt{n}$ is sufficient to recover the optimal
rate.
We use the \path{Nystroem} and \path{MiniBatchKmeans} classes from the \path{sklearn} python library \cite{scikit-learn}to implement kernel $k$-means with \nystrom embedding (\cref{alg:nyst-kk-means})
and we compute the solution $\bC_{n,m}^{++}$.

For our experiments we follow the same approach as \citet{wang2017scalable},
and test our algorithm on two variants of the MNIST digit dataset.
In particular, \mnistsmall \cite{lecun-mnisthandwrittendigit-2010}
is the original MNIST dataset containing pictures each with $d = 784$ pixels.
We divide each pixel by 255, bringing each feature in a $[0,1]$ interval.
We split the dataset in two part, $n = 60000$ samples are used to compute the $W(\bC_{n,m}^{++})$ centroids,
and we leave out unseen $10000$ samples to compute $W(\bC_{n,m}^{++},\mu_{test})$,
as a proxy for $W(\bC_{n,m}^{++}, \mu)$.
To test the scalability of our approach we also consider the \mnistbig dataset
from the infinite MNIST project \cite{loosli-canu-bottou-2006}, constructed using non-trivial transformations
and corruptions of the original \mnistsmall images. Here we compute $\bC_{n,m}^{++}$ using
$n = 8000000$ images, and compute $W(\bC_{n,m}^{++},\mu_{test})$ on $100000$ unseen images.
As in \citet{wang2017scalable} we use Gaussian kernel with bandwidth $\sigma = (1/n^2)\sqrt{\sum_{i,j} \normsmall{\bx_i - \bx_j}^2}$.

\begin{figure}[t]
\centering
\begin{tabular}{@{}cc@{}}
\includegraphics[width=0.49\textwidth]{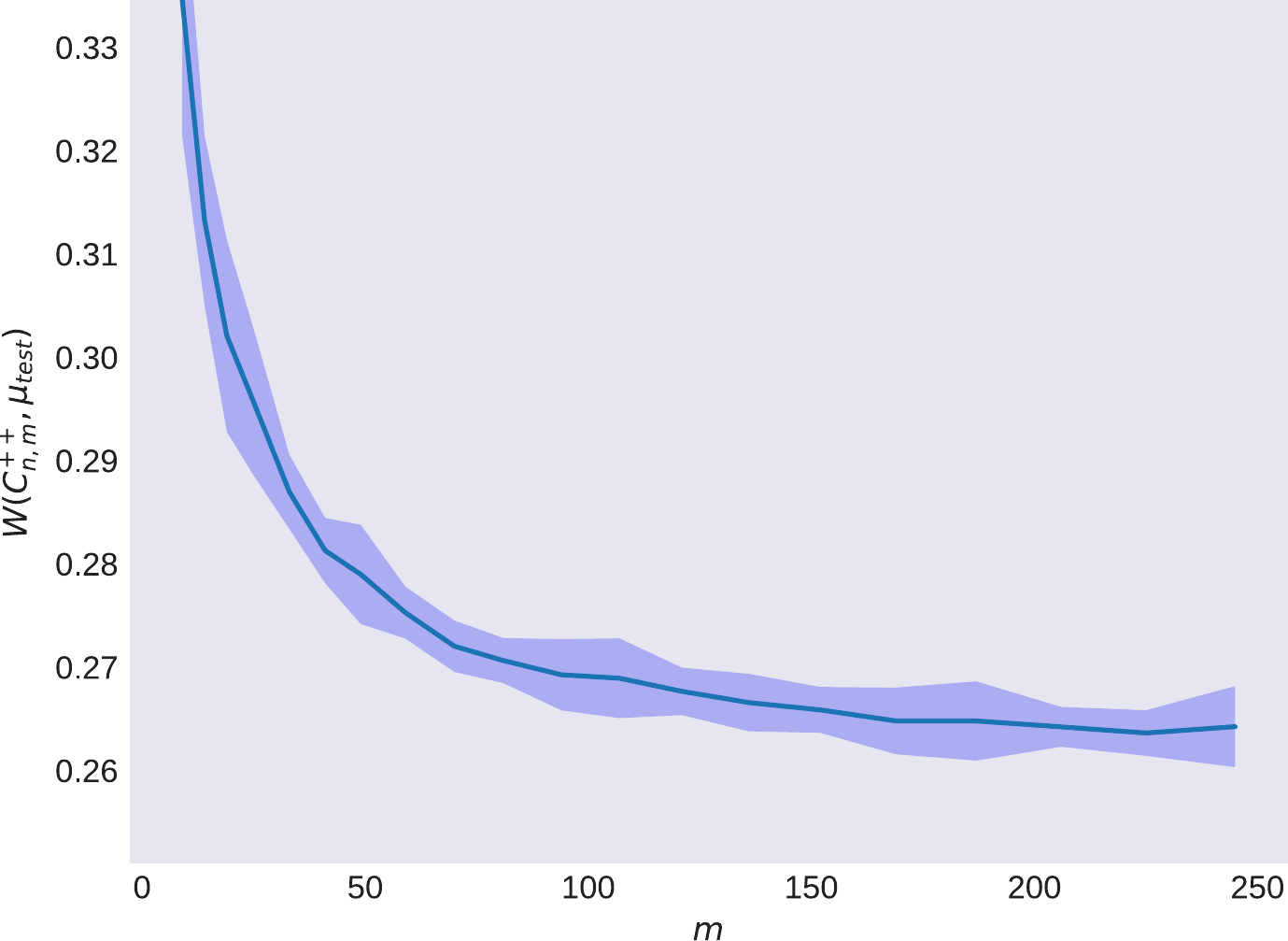}
&\includegraphics[width=0.49\textwidth]{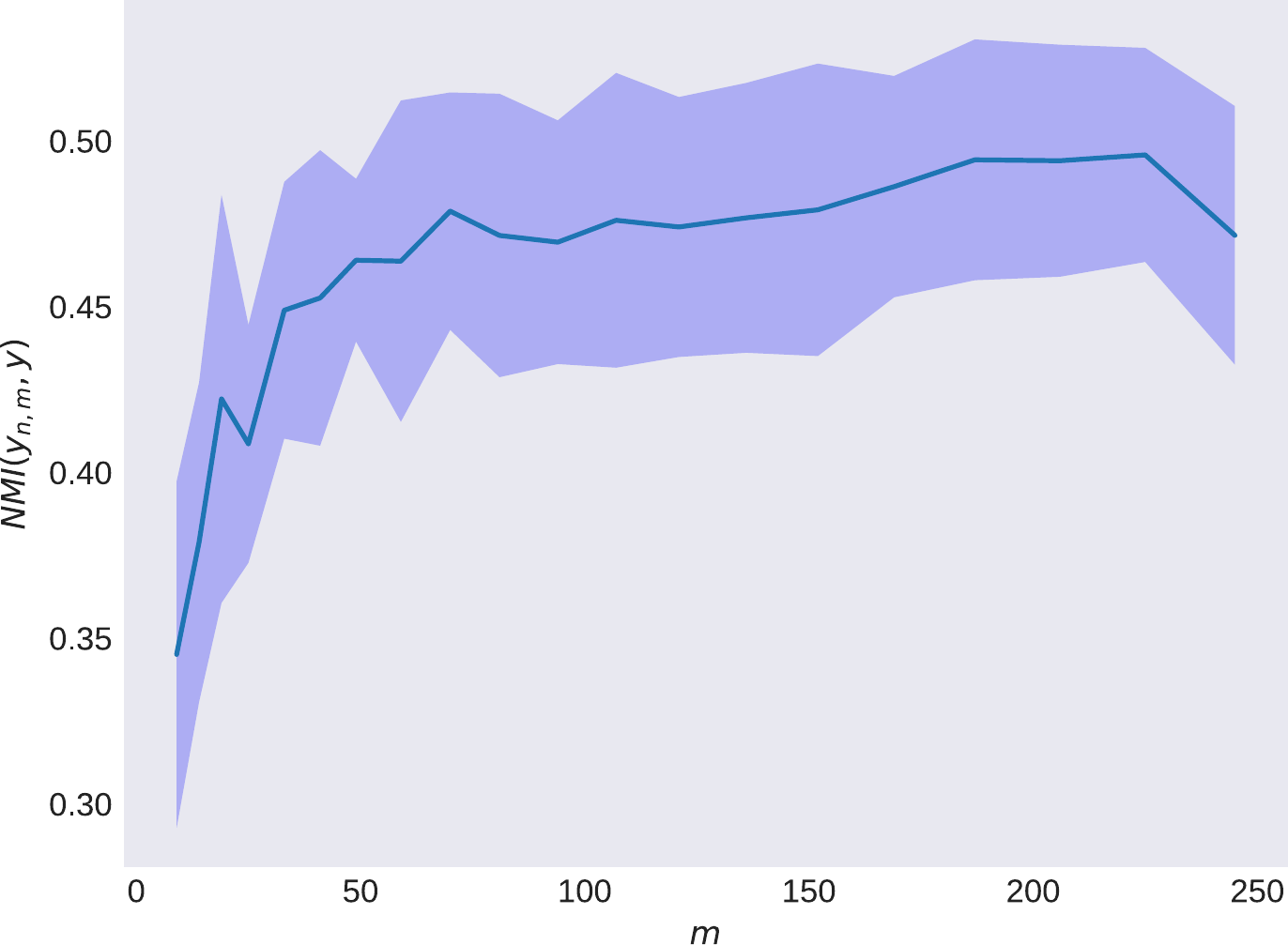}
\end{tabular}
\vspace{-.75\baselineskip}
\caption{Results for \mnistsmall}\label{fig:smallmnist}
\end{figure}
\textbf{\mnistsmall:} these experiments are small enough to run in less than a minute
on a single laptop with 4 cores and 8GB of RAM. The results are reported in
\cref{fig:smallmnist}. On the left we report in blue $W(\bC_{n,m}^{++}, \mu_{test})$,
where the shaded region is a $95\%$ confidence interval for the mean over 10 runs.
As predicted, the expected cost decreases as the
size of $\rkhs_m$ increases, and plateaus once we achieve $1/m \simeq 1/\sqrt{n}$,
in line with the statistical error.
Note that the normalized mutual information (NMI) between the true $[0-9]$ digit classes $\by$
and the computed cluster assignments $\by_{n,m}$ also plateaus around $1/\sqrt{n}$.
While this is not predicted by the theory, it strengthens the intuition
that beyond a certain capacity expanding $\rkhs_m$ is computationally wasteful.

\begin{figure}[t]
\centering
\begin{tabular}{@{}cc@{}}
\includegraphics[width=0.49\textwidth]{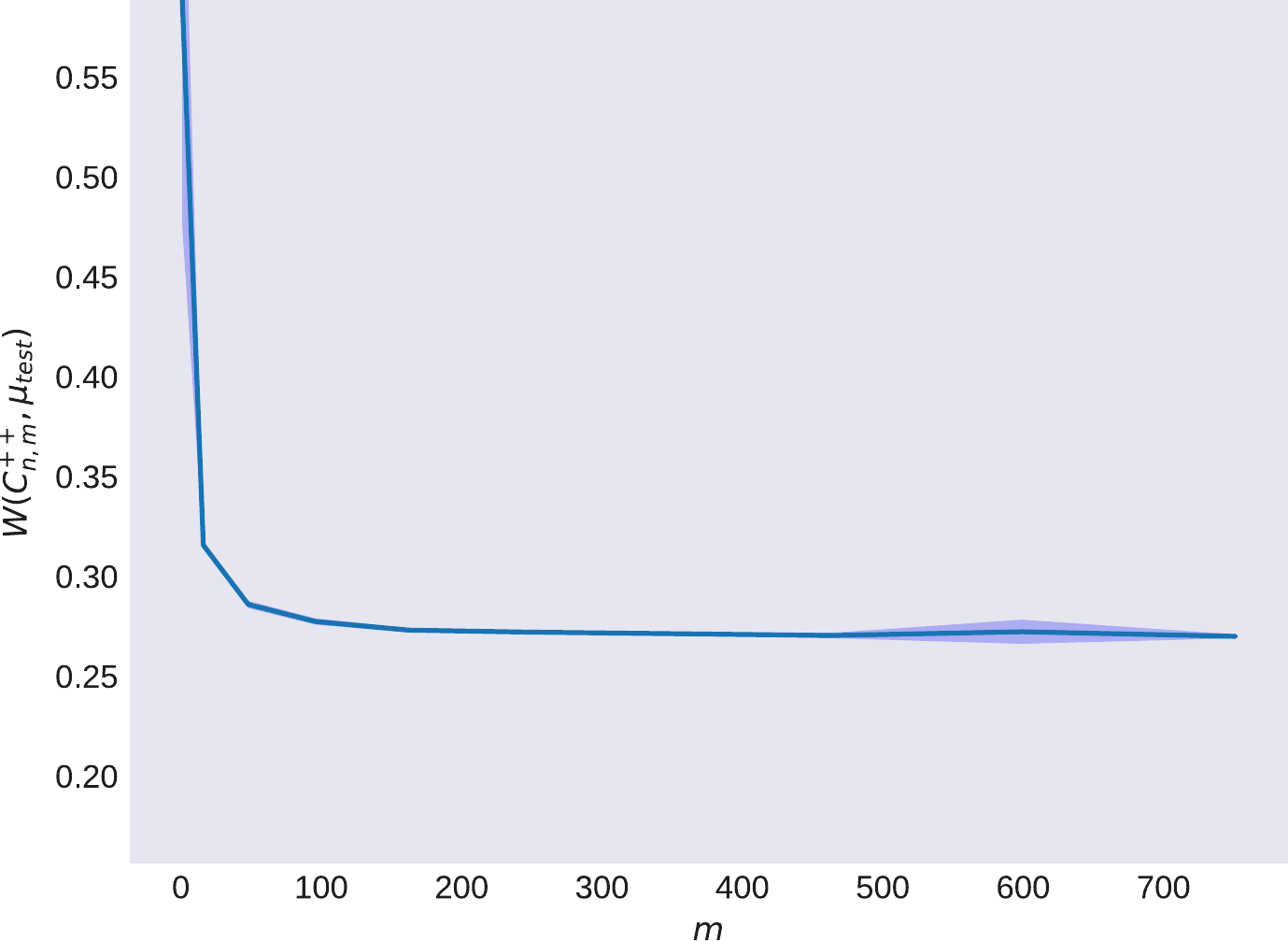}
&\includegraphics[width=0.49\textwidth]{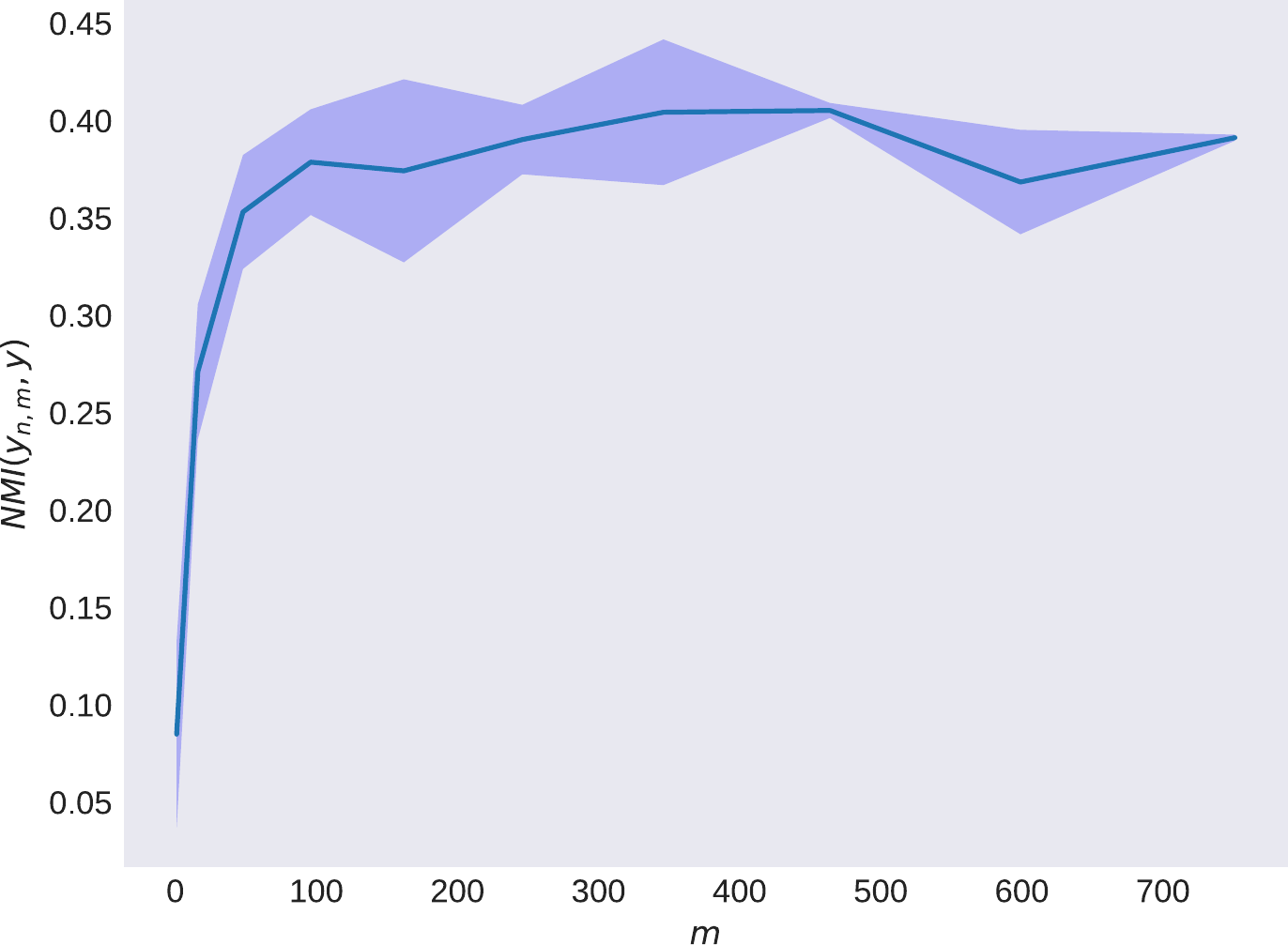}
\end{tabular}
\vspace{-.75\baselineskip}
\caption{Results for \mnistbig}\label{fig:bigmnist}
\end{figure}
\textbf{\mnistbig:} to test the scalability of our approach, we run the same experiment
on millions of points. Note that we carry out our \mnistbig experiment
on a \emph{single} 36 core machine with 128GB of RAM,
much less than the setup of \cite{wang2017scalable},
where at minimum a cluster of 8 such nodes are used.
The behaviour of $W(\bC_{n,m}^{++}, \mu_{test})$
and NMI are similar to \mnistsmall, with the increase in dataset size
allowing for stronger concentration and smaller confidence intervals.
Finalle, note that around $m = 400$ uniformly sampled landmarks are sufficient
to achieve $NMI(\by_{n,m}, \by) = 0.405$, matching the $0.406$ NMI reported by \cite{wang2017scalable}
for a larger $m = 1600$, although smaller than the $0.423$ NMI they report for
$m = 1600$ when using a slower, PCA based method to compute the embeddings,
and RLS sampling to select the landmarks. Nonetheless, computing $\bC_{n,m}^{++}$
takes less than 6 minutes on a single machine, while their best solution required
more than $1.5$hr on a cluster of 32 machines.
 \section{Open questions and conclusions}\label{sec:open-quest}
Combining \cref{lem:uniform-sampling-preserving} and \cref{lem:main-lemma},
we know that using uniform sampling we can linearly trade-off a $1/\gamma$
decrease in sub-space size $m$ with a $\gamma/n$ increase in excess risk.
While this is sufficient to maintain the $\bigotime(1/\sqrt{n})$ rate,
it is easy to see that the same would not
hold for a $\bigotime(1/n)$ rate, since we would need to
uniformly sample $n/1$ landmarks losing all computational improvements.

To achieve a better trade-off we must go beyond uniform sampling and
use different probabilities for each sample, to capture their uniqueness and contribution
to the approximation error.
\begin{definition}[\cite{alaoui2015fast}]\label{def:exact-ridge-lev-scores}
The $\gamma$-ridge leverage score (RLS) of point $i\in [n]$ is defined as
\begin{align}\label{eq:exact-rls}
    \tau_{i}(\gamma) = \phivec_i^\transp(\phimat_n\phimat_n^\transp + \gamma\pimat_n)^{-1}\phivec_i = \be_i^\transp\kermatrix_n(\kermatrix_n + \gamma\bI_n)^{-1}\be_i.
\end{align}
The sum of the RLSs
\begin{align*}
\deff^n(\gamma) = \sum_{i=1}^n \tau_{i}(\gamma) = \Tr\left(\kermatrix_n(\kermatrix_n + \gamma\bI_n)^{-1}\right) \leq \Rank(\kermatrix_n)/\gamma
\end{align*}
is the empirical effective dimension of the dataset.
\end{definition}
Ridge leverage scores are closely connected to the residual $\normsmall{\pimat_m^{\bot}\phivec_i}^2$ after the projection $\pimat_m$ discussed in \cref{def:regularized-nyst-app-guar}.
In particular, using \cref{lem:main-lemma} we have that the residual can be bounded as $\normsmall{\pimat_m^{\bot}\phivec_i}^2 \leq \tfrac{\gamma}{1-\varepsilon}\phivec_i^\transp(\phimat_n\phimat_n^\transp + \gamma\pimat_n)^{-1}\phivec_i$.
It is easy to see that, up to a factor $\tfrac{\gamma}{1-\varepsilon}$,
high-RLS points are also high-residual points. Therefore it is not surprising
that sampling according to RLSs quickly selects any high-residual points and covers $\rkhs_n$, generating a $\gamma$-preserving dictionary.
\begin{lemma}\cite{calandriello_thesis_2017} \label{lem:rls-sampling-preserving}
For a given $\gamma$, construct $\coldict$ by sampling
$m \geq 12\kappa^2\deff^n(\gamma)\log(n/\delta)/\varepsilon^2$
landmarks from $\dataset$ proportionally to their RLS.
Then w.p.\@ at least $1 - \delta$
the dictionary $\coldict$ is $\gamma$-preserving.
\end{lemma}
Note there exist datasets where the RLSs
are uniform,and therefore in the worst case the two sampling approaches coincide.
Nonetheless, when the data is more structured $m \simeq \deff^n(\gamma)$
can be much smaller than the $n/\gamma$ dictionary size required by uniform sampling.

Finally, note that computing RLSs exactly also requires constructing $\kermatrix_n$
and $\bigotime(n^2)$ time and space, but in recent years
a number of fast approximate RLSs sampling methods \cite{calandriello_disqueak_2017} have emerged
that can construct $\gamma$-preserving dictionaries of size $\abigotime(\deff^n(\gamma))$ in just
$\abigotime(n\deff^n(\gamma)^2)$ time.
Using this result, it is trivial to sharpen the computational aspects of
\cref{thm:main-result} in special cases.

In particular, we can generate a $\sqrt{n}$-preserving dictionary with only
$\deff^n(\sqrt{n})$ elements instead of the $\sqrt{n}$ required by uniform sampling.
Using concentration arguments \cite{rudi2015less}
we also know that \whp the empirical effective dimension
is at most three times $\deff^n(\gamma) \leq 3\deff^\sampdist(\gamma)$
the expected effective dimension, a $\sampdist$-dependent quantity that captures
the interaction between $\mu$ and the RKHS $\rkhs$.
\begin{definition}
Given the expected covariance operator $\bm{\Psi} := \expectedvalue_{\bx \sim \sampdist}\left[\phivec(\bx)\phivec(\bx)^\transp\right]$,
the \emph{expected}
effective dimension is defined as
$\deff^{\sampdist}(\gamma) = \expectedvalue_{\bx \sim \sampdist}\left[\phivec(\bx)\left(\bm{\Psi} + \gamma\pimat\right)^{-1}\phivec(\bx)\right].$
Moreover, for some constant $c$ that depends only on $\featmap(\cdot)$ and $\mu$,
$\deff^{\sampdist}(\gamma) \leq c\left(n/\gamma\right)^{\eta}$ with $0 < \eta \leq 1$.
\end{definition}
Note that $\eta = 1$ just gives us the $\deff^{\sampdist}(\gamma) \leq \bigotime(n/\gamma)$
worst-case upper bound that we saw for $\deff^n(\gamma)$, and it is always
satisfied when the kernel function is bounded.
If instead we have a faster spectral decay, $\eta$ can be much smaller.
For example, if the eigenvalues of $\bm{\Psi}$ decay polynomially as
$\lambda_i = i^{-\eta}$, then $\deff^{\sampdist}(\gamma) \leq c\left(n/\gamma\right)^{\eta}$,
and in our case $\gamma = \sqrt{n}$ we have $\deff^{\sampdist}(\sqrt{n}) \leq cn^{\eta/2}$.

We can now better characterize the gap between statistics and computation:
using RLSs sampling we can improve the computational aspect of \cref{thm:main-result}
from $\sqrt{n}$ to
$\deff^{\sampdist}(\gamma)$, but the risk rate remains
$\bigotime(k/\sqrt{n})$ due to the $\bigotime(k/\sqrt{n})$ component coming from \cref{prop:biau-risk-bound}.

Assume for a second we could generalize, with additional assumptions, \cref{prop:biau-risk-bound}
to a faster $\bigotime(1/n)$ rate. Then applying \cref{lem:main-lemma}
with $\gamma = 1$ we would obtain a risk $\risk(\bC_{n,m}) \leq \bigotime(k/n) + \bigotime(k/n)$.
Here we see how the regularity condition on $\deff^{\sampdist}(1)$ becomes crucial. In
particular, if $\eta = 1$, then we have $\deff^{\sampdist}(1) \sim n$ and no
gain. If instead $\eta < 1$ we obtain $\deff^{\sampdist}(1) \leq n^{\eta}$.
This kind of adaptive rates were shown to be possible in supervised learning
\cite{rudi2015less}, but seems to still be out of reach for approximate
kernel $k$-means.

One possible approach to fill this gap is to look at fast $\bigotime(1/n)$ excess risk
rates for kernel $k$-means.
\begin{proposition}[\cite{levrard2015nonasymptotic}, informal]\label{prop:fast-rate}
Assume that $k \geq 2$, and that $\sampdist$ satisfies a margin condition
with radius $r_0$.
If $\bC_n$ is an empirical risk minimizer, then, with probability larger than $1-e^{-\delta}$,
\begin{align*}
\risk(\bC_n) \leq \wt{\bigotime}\left(\frac{1}{r_0} \frac{(k + log (|M|))\log(1/\delta)}{n}\right),
\end{align*}
where $|M|$ is the cardinality of the set of all optimal (up to a relabeling) clustering.
\end{proposition}
For more details on the margin assumption, we refer the reader to the original paper \cite{levrard2015nonasymptotic}.
Intuitively the margin condition asks that every labeling (Voronoi grouping)
associated with an optimal clustering is reflected by large separation in $\sampdist$.
This margin condition also acts as a counterpart of the usual margin conditions for supervised
learning where $\sampdist$ must have lower density around
the neighborhood of the critical area
$\{ \bx | \mu'( Y = 1 | X = \bx) = 1 / 2\}$.
Unfortunately, it is not easy to integrate \cref{prop:fast-rate} in our analysis,
as it is not clear how the margin condition translate from $\rkhs_n$ to $\rkhs_m$.

\bibliographystyle{plainnat}

\newpage

\appendix
\section{Proofs}
\begin{proof}[Proof of \cref{thm:main-result}]
Given our dictionary $\coldict$, we decompose
\begin{align*}
\expectedvalue_{\dataset \sim \sampdist}[W(\bC_{n,m}, \sampdist)] - W^*(\sampdist)
&=\expectedvalue_{\dataset \sim \sampdist}[W(\bC_{n,m}, \sampdist) - W(\bC_{n}, \sampdist)] + \expectedvalue_{\dataset \sim \sampdist}[W(\bC_{n}, \sampdist)] - W^*(\sampdist)
\end{align*}
We can use \cref{prop:biau-risk-bound} to bound the second pair as
$\bigotime(k/\sqrt{n})$
using the proposition. Then we further split
\begin{align*}
W(\bC_{n,m}, \sampdist) - W(\bC_{n}, \sampdist)
&= W(\bC_{n,m}, \sampdist) - W(\bC_{n,m}, \sampdist_n)&& (a)\\
&+ W(\bC_{n,m}, \sampdist_n) - W(\bC_{n}, \sampdist_n)&& (b)\\
&+ W(\bC_{n}, \sampdist_n) - W(\bC_{n}, \sampdist)&&(c).
\end{align*}
The last line $(c)$ is negative, as $\bC_n$ is optimal \wrt $W(\cdot, \sampdist_n)$.
The first line $(a)$ can also be bounded by \cite[Lemma 4.3]{biau2008performance}, a stronger version of \cref{prop:biau-risk-bound}.
To bound the middle term $(b)$ we further expand
\begin{align*}
W(\bC_{n,m}, \sampdist_n) &= \frac{1}{n}\sum_{i=1}^n\min_{j=1,\dots,k}\normsmall{\phivec_i - \bc_{n,m,j}}^2.
\end{align*}
Note that $W$ evaluates the minimum $\min_{j=1,\dots,k}\normsmall{\phivec_i - \bc_{n,m,j}}^2$,
while $\bC_{n,m}$ has been constructed to minimize $\min_{j=1,\dots,k}\normsmall{\pimat_m\phivec_i - \bc_{n,m,j}}^2$. Nonetheless,
since $\bc_{n,m,j} \in \rkhs_m$ is orthogonal to $\pimat_m^{\bot}$
\begin{align*}
&\min_{j=1,\dots,k}\normsmall{\phivec_i - \bc_{n,m,j}}^2
= \min_{j=1,\dots,k}\normsmall{\pimat_m\phivec_i + \pimat_m^\bot\phivec_i - \bc_{n,m,j}}^2\\
&= \min_{j=1,\dots,k}\normsmall{\pimat_m\phivec_i - \bc_{n,m,j}}^2 + \normsmall{\pimat_m^\bot\phivec_i}^2 + \cancelto{0}{2(\pimat_m\phivec_i - \bc_{n,m,j})^\transp\pimat_m^\bot\phivec_i}\\
&= \min_{j=1,\dots,k}\normsmall{\pimat_m\phivec_i - \bc_{n,m,j}}^2 + \normsmall{\pimat_m^\bot\phivec_i}^2,
\end{align*}
and both criteria are minimized by the same $j$ (\ie they assign the point to the same cluster).

Before continuing, we must introduce additional notation to represent $\bC_n$ as the
average of the points in each cluster.
Let  $\{\bff_j\}_{j=1}^k$ be the cluster indicator vectors
such that $[\bff_j]_i = 1/|\cluster_j|$ if
sample $i$ is in the $j$-th cluster and $[\bff_j]_i = 0$ otherwise.
As a consequence $\norm{\bff_j}^2 = 1/|\cluster_j|$, and $[\bff_j]_i = 1/\norm{\bff_c}^2$.
Denote with $\bF \in \Real^{n \times k}$
the matrix containing $\bff_i$ as columns, and let $\feasibleset$ be the space
of feasible clustering, such that all $\bff_j$ are $\{0,1/\norm{\bff_c}^2\}$ binary,
and each row of $\bF$ contains only one non-zero entry. Let $\bR \in \Real^{k \times k}$ be the diagonal matrix
with $1/\norm{\bff_c}^2$ as the diagonal entries.

Then
$\bF^\transp\bF = \bR$ and $\bS := \bF\bR^{-1}\bF^\transp \in \Real^{n \times n}$ is a
projection matrix.
Finally we can express the centroids as $\bc_j = \phimat_n\bff_j = (1/|\cluster_j|)\sum \phivec_i$ and $\bC_n = \phimat_n\bF_n$.

Similarly, we define the optimal
$\bF_{n,m}$ associated with $\bC_{n,m}$,
of $\bR_{n,m} = \bF_{n,m}^\transp\bF_{n,m}$
and the projection matrix $\bS_{n,m} = \bF_{n,m}\bR_{n,m}^{-1}\bF_{n,m}^\transp$.
We still have that
\begin{align*}
\argmin_{j=1,\dots,k}\normsmall{\phivec_i - \bc_{n,m,j}}^2
= \pimat_m\phimat_n\bS_{n,m}\be_i,
\end{align*}
due to the optimality of $\bF_{n,m}$ \wrt $\pimat_m\phimat_n$. Substituting
in the definition of $W(\cdot,\cdot)$
\begin{align*}
W(\bC_{n,m},\sampdist_n)
&= \frac{1}{n}\sum_{i=1}^n\min_{j=1,\dots,k}\normsmall{\phivec_i - \bc_{n,m,j}}^2\\
&=\frac{1}{n}\sum_{i=1}^n\normsmall{\phivec_i - \pimat_m\phimat_n\bS_{n,m}\be_i}^2\\
&=\frac{1}{n}\normsmall{\phimat_n - \pimat_m\phimat_n\bS_{n,m}}_F^2\\
&= \frac{1}{n}\Tr(\phimat_n^\transp\phimat_n - 2\phimat_n^\transp\pimat_m\phimat_n\bS_{n,m}
+ \bS_{n,m}\phimat_n^\transp\pimat_m\pimat_m\phimat_n\bS_{n,m}).
\end{align*}
Since $\bS_{n,m}$ is a projection matrix
\begin{align*}
\Tr(\phimat_n^\transp\pimat_m\phimat_n\bS_{n,m})
&= \Tr(\phimat_n^\transp\pimat_m\phimat_n\bS_{n,m}\bS_{n,m})\\
&= \Tr(\bS_{n,m}\phimat_n^\transp\pimat_m\phimat_n\bS_{n,m}),
\end{align*}
and using the optimality of $\bF_{n,m}$ we have
\begin{align*}
&\normsmall{\phimat_n - \pimat_m\phimat_n\bS_{n,m}}_F^2
= \Tr(\phimat_n^\transp\phimat_n - \bS_{n,m}\phimat_n^\transp\pimat_m\phimat_n\bS_{n,m})\\
&= \Tr(\phimat_n^\transp\pimat_m\phimat_n - \bS_{n,m}\phimat_n^\transp\pimat_m\phimat_n\bS_{n,m}) + \Tr(\phimat_n^\transp\phimat_n - \phimat_n^\transp\pimat_m\phimat_n)\\
&\leq  \Tr(\phimat_n^\transp\pimat_m\phimat_n - \bS_{n}\phimat_n^\transp\pimat_m\phimat_n\bS_{n}) + \Tr(\phimat_n^\transp\phimat_n - \phimat_n^\transp\pimat_m\phimat_n)\\
&= \Tr(\phimat_n^\transp\phimat_n - \bS_{n}\phimat_n^\transp\pimat_m\phimat_n\bS_{n}).
\end{align*}
Using \cref{def:regularized-nyst-app-guar} we have $\pimat_m - \pimat_n \succeq  -\frac{\gamma}{1-\varepsilon}(\phimat_n\phimat_n^\transp + \gamma\pimat_n)^{-1}$ and
\begin{align*}
&\Tr(\phimat_n^\transp\phimat_n - \bS_{n}\phimat_n^\transp\pimat_m\phimat_n\bS_{n})\\
&\leq \Tr\left(\phimat_n^\transp\phimat_n - \bS_{n}\phimat_n^\transp\phimat_n\bS_{n}
+ \frac{\gamma}{1-\varepsilon}\bS_{n}\phimat_n^\transp(\phimat_n\phimat_n^\transp + \gamma\pimat_n)^{-1}\phimat_n\bS_{n}\right)\\
&\leq \Tr\left(\phimat_n^\transp\phimat_n - \bS_{n}\phimat_n^\transp\phimat_n\bS_{n}
+ \frac{\gamma}{1-\varepsilon}\bS_{n}\phimat_n^\transp(\phimat_n\phimat_n^\transp)^{+}\phimat_n\bS_{n}\right)\\
&= W(\bC_n,\sampdist_n)
+ \frac{\gamma}{1-\varepsilon}\Tr\left(\bS_{n}\phimat_n^\transp(\phimat_n\phimat_n^\transp)^+\phimat_n\bS_{n}\right).
\end{align*}
Noting now that $\normsmall{\phimat_n^\transp(\phimat_n\phimat_n^\transp)^+\phimat_n} \leq 1$,
$\bS_n$ is a projection matrix, and $\Tr(\bS_{n}) = k$ we have
\begin{align*}
\frac{\gamma}{1-\varepsilon}\Tr\left(\bS_{n}\phimat_n^\transp(\phimat_n\phimat_n^\transp)^+\phimat_n\bS_{n}\right)
\leq \frac{\gamma}{1-\varepsilon}\Tr\left(\bS_{n}\bS_{n}\right)
=  \frac{\gamma k}{1-\varepsilon}.
\end{align*}
Conversely, if we focus on the projection matrix $\phimat_n^\transp(\phimat_n\phimat_n^\transp)^+\phimat_n = \phimat_n^\transp\phimat_n(\phimat_n\phimat_n^\transp)^+ = \kermatrix_n\kermatrix_n^{+}$, and letting $r = \Rank(\kermatrix_n)$ we have
\begin{align*}
\frac{\gamma}{1-\varepsilon}\Tr\left(\bS_{n}\phimat_n^\transp(\phimat_n\phimat_n^\transp)^+\phimat_n\bS_{n}\right)
\leq \frac{\gamma}{1-\varepsilon}\Tr\left(\phimat_n^\transp(\phimat_n\phimat_n^\transp)^+\phimat_n\right)
\leq  \frac{\gamma r}{1-\varepsilon}.
\end{align*}
Since both bounds hold simultaneously, we can simply take the minimum to conclude
our proof.
\end{proof}

\begin{proof}[Proof of \cref{thm:main-kpp-result}]
Given our dictionary $\coldict$, we need to change our decomposition. We
denote with $\expectedvalue_{\mathcal{A}}[W(\bC_{n,m}^{++},\mu)]$ the expectation
over the randomness of the $k$-means++ seeding and Loyd algorithm.
Then
\begin{align*}
\expectedvalue_{\dataset \sim \sampdist}\left[\expectedvalue_{\mathcal{A}}[W(\bC_{n,m}^{++}, \sampdist)]\right]
= \expectedvalue_{\dataset \sim \sampdist}\left[\expectedvalue_{\mathcal{A}}[W(\bC_{n,m}^{++}, \sampdist)] - W(\bC_{n,m}^{++},\sampdist_n)\right]
+ \expectedvalue_{\dataset \sim \sampdist}\left[\expectedvalue_{\mathcal{A}}[W(\bC_{n,m}^{++},\sampdist_n)\right].
\end{align*}
Once again the first term can be bounded as $\bigotime(k/\sqrt{n})$
using the stronger \cite[Lemma 4.3]{biau2008performance}, so we turn our attention on the second term.
From \cref{prop:kmpp-opt-lemma} we have
\begin{align*}
\expectedvalue_{\dataset \sim \sampdist}\left[\expectedvalue_{\mathcal{A}}[W(\bC_{n,m}^{++},\sampdist_n)\right]
&\leq \expectedvalue_{\dataset \sim \sampdist}\left[8(\log(k) + 2)W(\bC_{n,m},\sampdist_n)\right]\\
&\leq \expectedvalue_{\dataset \sim \sampdist}\left[8(\log(k) + 2)\left(W(\bC_{n},\sampdist_n) + \frac{k}{1-\varepsilon}\frac{\gamma}{n}\right)\right],
\end{align*}
where we used \cref{lem:rls-sampling-preserving} in the second inequality.
Adding and subtracting $W(\bC_m,\,u)$, using once again \cref{prop:biau-risk-bound}, and putting everything together we have
\begin{align*}
\expectedvalue_{\dataset \sim \sampdist}\left[\expectedvalue_{\mathcal{A}}[W(\bC_{n,m}^{++}, \sampdist)]\right]
&\leq \bigotime\left(\frac{k}{\sqrt{n}}\right)
+ \expectedvalue_{\dataset \sim \sampdist}\left[\expectedvalue_{\mathcal{A}}[W(\bC_{n,m}^{++},\sampdist_n)\right]\\
&\leq \bigotime\left(\frac{k}{\sqrt{n}}\right)
+ 8(\log(k) + 2)\left(\expectedvalue_{\dataset \sim \sampdist}\left[W(\bC_{n},\sampdist_n)\right] + \frac{k}{1-\varepsilon}\frac{\gamma}{n}\right)\\
&\leq \bigotime\left(\frac{k}{\sqrt{n}}\right)
+ 8(\log(k) + 2)\left(\bigotime\left(\frac{k}{\sqrt{n}}\right) + W^{*}(\sampdist) + \frac{k}{1-\varepsilon}\frac{\gamma}{n}\right)\\
&\leq 
\bigotime\left(\log(k)\left(\frac{k}{\sqrt{n}} + W^{*}(\sampdist) + k\frac{\gamma}{n}\right)\right),
\end{align*}
which concludes our proof.
\end{proof}

\begin{proof}[Proof of \cref{lem:main-lemma}]
Before starting the proof, we need the following result, which is a trivial extension
of \cite[Lemma 2]{bach2013sharp} to an RKHS, see also \cite{calandriello_disqueak_2017}.
\begin{corollary}[\cite{bach2013sharp}]\label{cor:bach-conc}
Let $\coldict$ be constructed by uniformly sampling $m \geq 12\kappa^2n/\gamma\log(n/\delta)/\varepsilon^2$ points from $\phimat_n$ with replacement.
Then with probability at least $1 - \delta$ we have
\begin{align*}
\norm{\left(\phimat_n\phimat_n^\transp + \gamma\pimat_n\right)^{-1/2}\left(\phimat_n\phimat_n^\transp - \frac{n}{m}\phimat_m\phimat_m^\transp\right)\left(\phimat_n\phimat_n^\transp + \gamma\pimat_n\right)^{-1/2}} \leq \varepsilon,
\end{align*}
which implies
\begin{align*}
(1-\varepsilon)\phimat_n\phimat_n^\transp - \varepsilon\gamma\pimat_n \preceq \frac{n}{m}\phimat_m\phimat_m^\transp \preceq (1+\varepsilon)\phimat_n\phimat_n^\transp + \varepsilon\gamma\pimat_n
\end{align*}
\end{corollary}
Using $\pimat_n$'s definition
\begin{align*}
\pimat_n &= \left(\frac{n}{m}\phimat_m\phimat_m^\transp + \varepsilon\gamma\pimat_n\right)^{-1/2}\left(\frac{n}{m}\phimat_m\phimat_m^\transp + \varepsilon\gamma\pimat_n\right)\left(\frac{n}{m}\phimat_m\phimat_m^\transp + \varepsilon\gamma\pimat_n\right)^{-1/2}\\
&\begin{aligned}
= \left(\frac{n}{m}\phimat_m\phimat_m^\transp + \varepsilon\gamma\pimat_n\right)^{-1/2}&\frac{n}{m}\phimat_m\phimat_m^\transp\left(\frac{n}{m}\phimat_m\phimat_m^\transp + \varepsilon\gamma\pimat_n\right)^{-1/2} \\
&+ \varepsilon\gamma\left(\frac{n}{m}\phimat_m\phimat_m^\transp + \varepsilon\gamma\pimat_n\right)^{-1/2}\pimat\left(\frac{n}{m}\phimat_m\phimat_m^\transp + \varepsilon\gamma\pimat_n\right)^{-1/2}
\end{aligned}\\
&= \left(\frac{n}{m}\phimat_m\phimat_m^\transp + \varepsilon\gamma\pimat_n\right)^{-1/2}\frac{n}{m}\phimat_m\phimat_m^\transp\left(\frac{n}{m}\phimat_m\phimat_m^\transp + \varepsilon\gamma\pimat_n\right)^{-1/2} + \varepsilon\gamma\left(\frac{n}{m}\phimat_m\phimat_m^\transp + \varepsilon\gamma\pimat_n\right)^{-1}\\
&= \left(\frac{n}{m}\phimat_m\phimat_m^\transp + \varepsilon\gamma\pimat_n\right)^{-1/2}\frac{n}{m}\phimat_m\phimat_m^\transp\left(\frac{n}{m}\phimat_m\phimat_m^\transp + \varepsilon\gamma\pimat_n\right)^{-1/2} + \varepsilon\gamma\left(\frac{n}{m}\phimat_m\phimat_m^\transp + \varepsilon\gamma\pimat_n\right)^{-1}.
\end{align*}
Note now that
\begin{align*}
&\left(\frac{n}{m}\phimat_m\phimat_m^\transp + \varepsilon\gamma\pimat_n\right)^{-1/2}\frac{n}{m}\phimat_m\phimat_m^\transp\left(\frac{n}{m}\phimat_m\phimat_m^\transp + \varepsilon\gamma\pimat_n\right)^{-1/2}\\
&= \frac{n}{m}\phimat_m\left(\frac{n}{m}\phimat_m\phimat_m^\transp + \varepsilon\gamma\pimat_n\right)^{-1}\phimat_m^\transp
= \phimat_m\left(\phimat_m\phimat_m^\transp + \frac{m}{n}\varepsilon\gamma\pimat_n\right)^{-1}\phimat_m^\transp\\
&\preceq \phimat_m\left(\phimat_m\phimat_m^\transp\right)^{+}\phimat_m^\transp
= \pimat_m
\end{align*}
And that using \cref{cor:bach-conc} we have
\begin{align*}
\left(\frac{n}{m}\phimat_m\phimat_m^\transp + \gamma\pimat_n\right)^{-1}
&\preceq (\left(1-\varepsilon\right)\phimat_n\phimat_n^\transp + (\varepsilon\gamma-\varepsilon\gamma)\pimat_n)^{-1}\\
&= \frac{1}{1-\varepsilon}\left(\phimat_n\phimat_n^\transp + \frac{\varepsilon\gamma - \varepsilon\gamma}{1-\varepsilon}\pimat_n\right)^{-1}
= \frac{\gamma}{1-\varepsilon}\left(\phimat_n\phimat_n^\transp\right)^{+}.
\end{align*}
Combining these two results we obtain the proof.
\end{proof}
 
\end{document}